\documentclass{bmvc2k}

\usepackage{times}
\usepackage{epsfig}
\usepackage{graphicx}
\usepackage{amsmath}
\usepackage{mathtools}
\usepackage{microtype}
\usepackage{amssymb}
\usepackage{subfigure}

\usepackage{algorithm}
\usepackage{algpseudocode}
\usepackage{booktabs}
\usepackage{multirow}
\usepackage{xcolor}
\usepackage{color, colortbl}
\usepackage{pifont}
\usepackage{capt-of}
\usepackage[normalem]{ulem}
\usepackage[toc,page]{appendix}
\usepackage[symbol]{footmisc}

\makeatletter
\newcommand{\printfnsymbol}[1]{%
  \textsuperscript{\@fnsymbol{#1}}%
}
\makeatother

\definecolor{m_purple}{HTML}{6F2DBD}
\definecolor{m_green}{HTML}{2C8915}
\definecolor{r_impro}{HTML}{6AAA96}
\definecolor{r_impro_n}{HTML}{E67F83}
\definecolor{algo_comments}{HTML}{297929}

\newcommand{\checkmarkgreen}{\textcolor{m_green}{\checkmark}}%
\newcommand{\xmarkred}{\textcolor{red}{\ding{55}}}%
\newcommand{\algrule}[1][.2pt]{\par\vskip.3\baselineskip\hrule height #1\par\vskip.3\baselineskip}

\title{Self-Supervised Learning in Multi-Task Graphs through Iterative Consensus Shift}

\addauthor{Emanuela Haller$^{* \text{1,}}$}{haller.emanuela@gmail.com}{2}
\addauthor{Elena Burceanu$^{* \text{1,}}$} {eburceanu@bitdefender.com}{3}
\addauthor{Marius Leordeanu$^{\text{1,2,}}$}{leordeanu@gmail.com}{4}

\addinstitution{
 Bitdefender, Romania
}
\addinstitution{
 University Politehnica of Bucharest,\\ Romania
}
\addinstitution{
 University of Bucharest, Romania
}
\addinstitution{
 Institute of Mathematics of the\\Romanian Academy, Romania
}

\runninghead{E. Haller, E. Burceanu, M. Leordeanu}{Consensus Shift}

\def\eg{\emph{e.g}\bmvaOneDot}

\begin{document}

\maketitle

\begin{abstract}
The human ability to synchronize the feedback from all their senses inspired recent works in multi-task and multi-modal learning. While these works rely on expensive supervision, our multi-task graph requires only pseudo-labels from expert models. Every graph node represents a task, and each edge learns between tasks transformations. Once initialized, the graph learns self-supervised, based on a novel consensus shift algorithm that intelligently exploits the agreement between graph pathways to generate new pseudo-labels for the next learning cycle. 
We demonstrate significant improvement from one unsupervised learning iteration to the next, outperforming related recent methods in extensive multi-task learning experiments on two challenging datasets. Our code is available at \url{https://github.com/bit-ml/cshift}.
\end{abstract}

\section{Introduction}

\footnotetext{$^*$ Equal contribution.}
Seeing the world from multiple perspectives offers a rich source of knowledge, as recent works show~\cite{active-learning-ensemble, experts-az, multi-features, progres-ensemble, multi-task-loss, taskonomy}. While multi-tasks methods attain a comprehensive understanding of the scene, they require a larger amount of supervision than single-task ones. Different from multi-modal~\cite{mmmt_loss, gdt} and multi-task graph approaches \cite{ngc, xtc}, we overcome the expensive labeling problem in two steps, taking advantage of existing experts in the literature, pretrained for different tasks. We use them to initially train a multi-task graph, where every node represents a task, and each edge transforms one task into another. We use the mutual consensus along different paths reaching a given task as a self-supervisory signal to further improve over the initial experts. We differ from the most related works~\cite{ngc, xtc} in two important ways: 
1) we show that without access to labeled data for the target domain, we can initialize the graph using experts from other domains and significantly improve over their performance; 2) our intelligent consensus-finding selection procedure, CShift, which adaptively considers the importance of each incoming edge to a node, is significantly more effective than simple averaging with fixed graph structure~\cite{ngc}. Our \textbf{key contributions} are:

    1. \textbf{We present iterative Consensus Shift (CShift)}, a method for unsupervised multi-task learning for new, unseen data distributions. CShift exploits, with an adaptive edge selection procedure, the consensus among multiple pathways reaching a given node (task), which becomes a supervisory signal at that node. Learning continues over multiple iterations, during which node pseudo-labels shift their values and improve accuracy after each iteration. While initial labels are provided by experts pre-trained on other datasets, the multi-task graph successfully adapts to the new data distribution.

    2. \textbf{We validate our claims in extensive experiments} on two recent datasets. CShift self-improves in an unsupervised manner, over multiple tasks, from one iteration to the next, while significantly outperforming the state-of-the-art experts used for the initial pseudo-labels.

\begingroup
\setlength{\tabcolsep}{2pt} 
\begin{table}[t]
    \begin{center}
        \begin{tabular}{l cccccc}
        \toprule
        & 
        \shortstack{Supervision on\\target domain} & 
        \shortstack{Uses\\ensembles}  &
        \shortstack{Ensemble as\\supervision}  &
        \shortstack{Selection\\in ensemble} &
        \shortstack{Unsupervised\\domain adapt.} & 
        \shortstack{Full\\graph}\\
        \midrule
        XTC~\cite{xtc} & supervised  & \xmarkred & N/A & N/A & \xmarkred & \checkmarkgreen \\
        
        NGC~\cite{ngc} & semi-supervised & \checkmarkgreen & \checkmarkgreen & \xmarkred  & \xmarkred & \xmarkred \\
        
        \textbf{CShift} (ours) & unsupervised & \checkmarkgreen & \checkmarkgreen & \checkmarkgreen & \checkmarkgreen & \checkmarkgreen  \\
        \bottomrule
        \end{tabular}
    \end{center}
    \setlength{\abovecaptionskip}{-12pt}
    \setlength{\belowcaptionskip}{-12pt}
    \caption{Learning in Multi-Task Graphs. CShift algorithm vs State-of-the-Art methods.}
    \label{tab:compare_features}
\end{table}
\endgroup

\subsection{Relation to prior work}\label{sec:related_work}

\noindent\textbf{Relation to Ensembles and Experts.} The idea of many paths working together to reach a common goal was often demonstrated~\cite{ensemble_survey} over time. We guide learning using a set of expert models, which has proven effective on video and image retrieval~\cite{experts_cordelia, experts-az, experts_video_retrieval, experts_videotext_retrieval}.

\noindent\textbf{Relation to Unsupervised Representation Learning.} Recent works use pretext tasks \cite{predict-rotation, split-brain, shuffle-herbert, colorization}, perform clustering \cite{deepcluster-v2, unsup-clust-cvpr, cluster-herbert}, minimize contrastive noise \cite{noise-contrast, contr-noise-ziss, constrast-cord-isola} or train adversarial generative models \cite{bigbigan}. Different from them, we train on pseudo-ground truth, constructed with our CShift algorithm, from multiple graph paths reaching the same task.

\noindent\textbf{Relation to Multi-Modal Learning.}
Recent papers combine modalities and tasks~\cite{mmmt_loss}, often using using multi-modal data transformation as self-supervision~\cite{gdt, cmc, contr-noise-ziss}. Different from them, we learn complex interactions between tasks without ground truth by modeling them with a bidirectional multi-task graph, using multi-path consensus and selection as supervision.

\noindent\textbf{Relation to Unsupervised Domain Adaptation (UDA).} A common UDA approach is to reduce discrepancy in feature space~\cite{uda_featspace_1_tzeng2014deep, uda_featspace_2_long2015learning}. Others use adversarial training with semantic and style consistency \cite{r3_2_hoffman2016fcns, r3_1_hoffman2018cycada}. Ours directly transforms between multiple tasks and exploits across task pixel-level consistencies. The method in \cite{atdt_ramirez2019learning} learns across two tasks in a particular setup with supervision for both in the source domain and only for one in the target domain. An extension with adversarial training is introduced in \cite{r3_3_chavhan2021ada}. In our case, we do not require annotation for the target domain and rely entirely on consensus between multiple tasks.

\begin{figure}[t!]
    \begin{center}
    \includegraphics[width=1\textwidth]{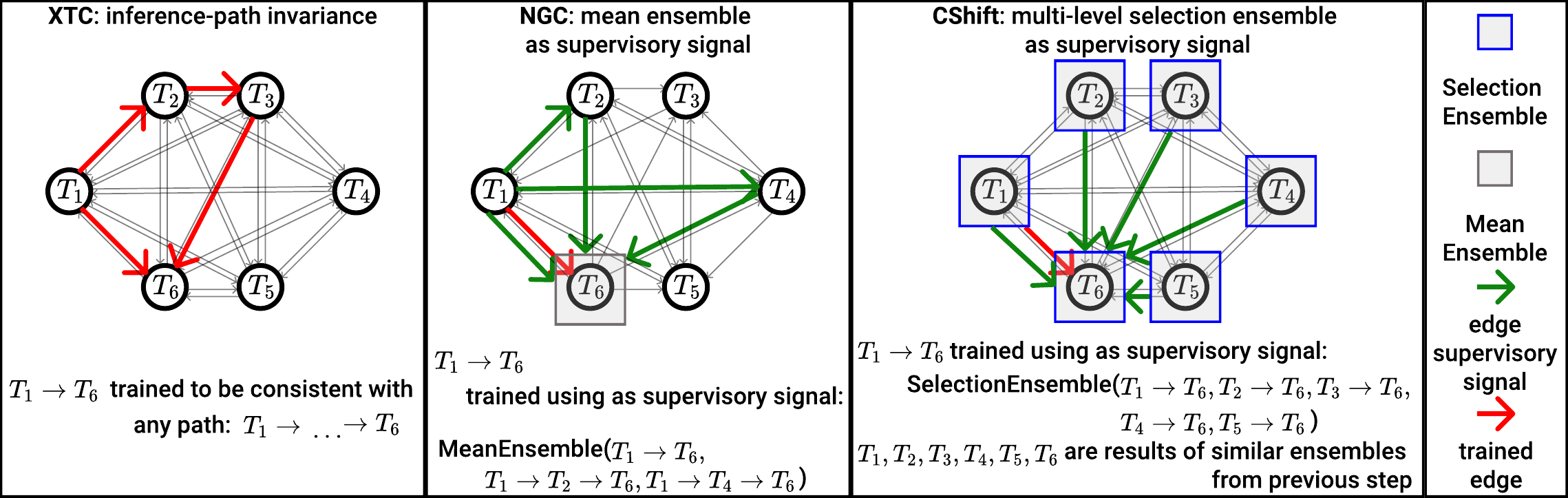} 
    \end{center}
    \setlength{\abovecaptionskip}{-20pt}
    \caption{Training strategies employed by other Multi-Task Graph methods compared with CShift. Different from NGC, our selection is unsupervised and uses the consensus of all the graph's edges. Node views are replaced with the consensus of the in-edges (blue box) and ensemble results become both inputs of out-edges and supervisory signals in the next iterations, generating a faster knowledge propagation.}\label{fig:xtc_ngc_cshift}
    
\end{figure}

\noindent\textbf{Relation to Multi-Task Learning.} 
 \cite{taskonomy, xtc} shows that underlying connections between different tasks can be exploited to effectively reduce the labeled data required for training. We, however, assume no multi-task annotated data on the target domain but only rely on per-task expert models, pre-trained on different domains, for initialization. After initialization, our CShift learns, in fact, completely unsupervised on the target domain. From the architectural point of view, our model is related to the recent Neural Graph Consensus (NGC) model~\cite{ngc}, which also connects multiple interpretations and tasks into a single graph of neural networks. 
Our model differs from NGC in four essential aspects:
\textbf{1)} the proposed selection mechanism is highly adaptive (different for each pixel, sample, and iteration), allowing a dynamical adjustment of the graph structure as detailed in Sec.~\ref{sec:approach_consensus_shift} and Fig.~\ref{fig:novelty_visual}, compared with the simple average in NGC; \textbf{2)} our fully connected graph guides the learning process only based on unsupervised consensus, as opposed to having a fixed, pruned architecture based on supervised data (NGC); \textbf{3)} we use the ensemble labels both as a supervisory signal at a node and as input for all its out-edges, making learning more efficient; \textbf{4)} we initialize the graph with pseudo-labels generated by out-of-distribution experts, while
NGC assumes a fully supervised initialization.
 In Tab.~\ref{tab:compare_features} and Fig.~\ref{fig:xtc_ngc_cshift} we show key differences against XTC~\cite{xtc} and NGC~\cite{ngc}.

\section{Our Approach}\label{sec:approach}

\begin{figure}[t!]
    \begin{center}
    \includegraphics[width=0.95\textwidth]{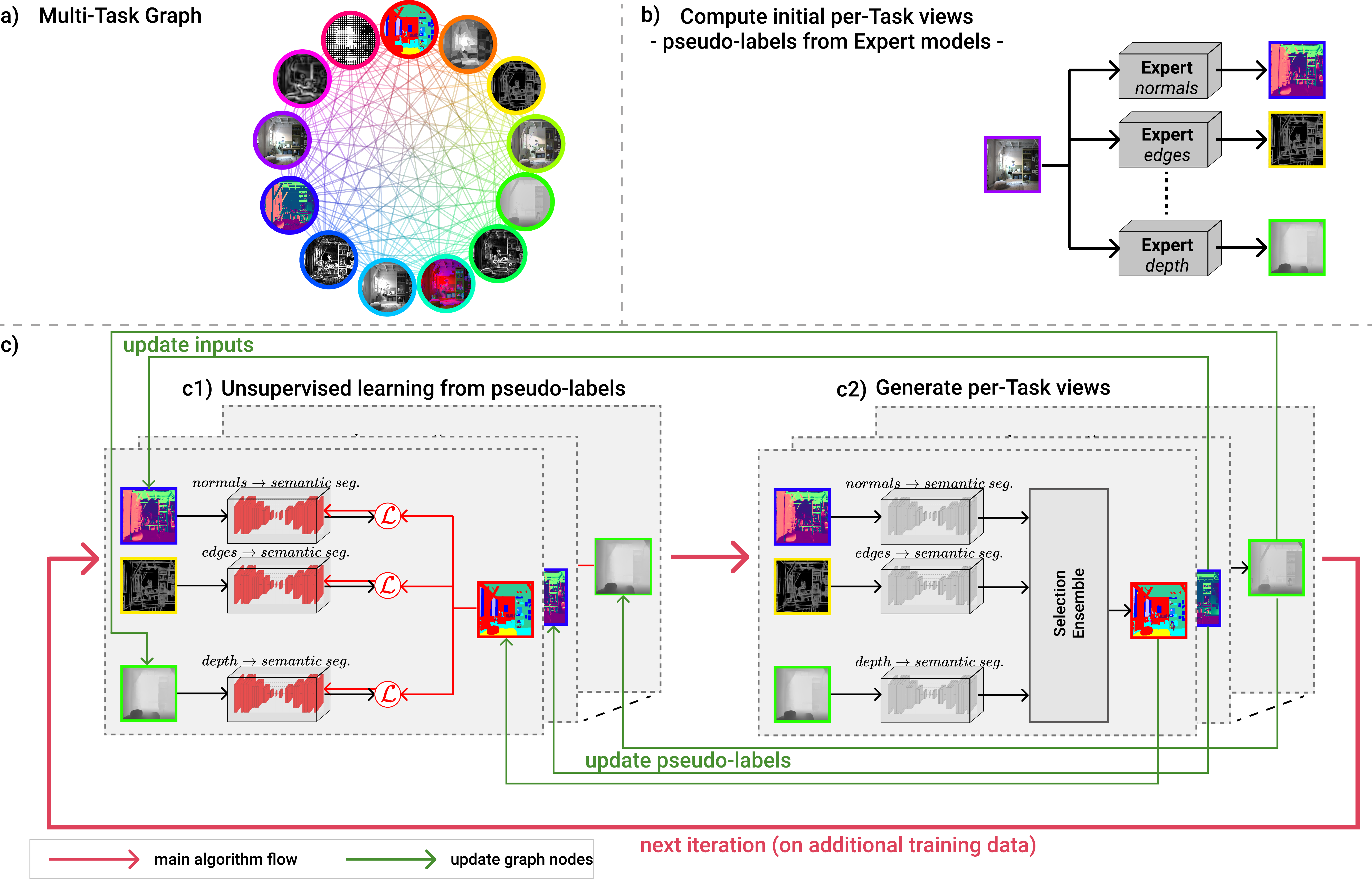}
    \end{center}
    \setlength{\abovecaptionskip}{-15pt}
    \caption{\textbf{CShift} architecture. \textbf{a) Fully-connected Multi-Task Graph}, with $13$ nodes (tasks). Edges are neural nets, transforming source to destination tasks. \textbf{b) Initialization from experts.} Based on the $rgb$ image, experts (black boxes in our system, trained on different distributions than ours) predict the initial pseudo-labels for each task. \textbf{c) CShift's iterations.} We train each graph edge using pseudo-labels. For each node, we compute its new pseudo-labels as the consensual representations of its in-edges by the intelligent CShift ensemble mechanism, which adaptively changes (differently for each pixel and data sample) the importance of each in-edge in the ensemble. The newly computed labels of a node become its supervisory signal in the next iteration and also inputs for all the out-edges of the node.}
    \label{fig:main_alg}
\end{figure}

We propose a novel Multi-Task Graph (Fig.~\ref{fig:main_alg}-a), which uses as supervision the consensual output, extracted through an intelligent CShift selection procedure (Fig.~\ref{fig:main_alg}-c), over multiple graph pathways that reach a given node. As previously mentioned, each graph node represents a task (or a view of the world). Each edge is a neural net that transforms a task at one node into another at a different node. Our graph is directed and fully connected.
In Fig.~\ref{fig:main_alg} we illustrate the main steps of our approach. All edges are initially trained using node pseudo-labels generated by out-of-distribution experts. After initialization (Fig.~\ref{fig:main_alg}-b), we set up the \textbf{view associated with a node}, computed as the CShift ensemble result of all the node's in-edges. These views become the pseudo-ground truth labels during the subsequent learning iterations in the graph. The views (pseudo-label values) shift from one learning iteration to the next, according to the CShift algorithm.
CShift uses the views at each node to transmit information through the out-edges towards other nodes and collects the in-edges' information to create the new views, at the next iteration, by a selection mechanism that establishes the consensus over the multiple incoming edges. The process is repeated until equilibrium is found at convergence. The iterative learning phase is unsupervised and initial experts could be created independently, using information from other datasets and domains, as our tests show.

\setlength{\textfloatsep}{0.3cm}
\begin{algorithm}
    \caption{- \textbf{CShift: Multi-Task Graph Learning with Consensus Shift}}
    \label{alg:main_alg}
        $X_{i}$ \;\;\;\;\;\;\;\;\;\;\,\;\;\;\;\;\;\;\;\; - input data sample $i$\;\;\;\;\;\;\;\;\;\;\;\;\;\;\;\;\;\, 
        $Y_{i;d}$ \;\, - pseudo-label for data $i$, task $d$\\
        Expert$_d$ \;\;\;\;\;\;\;\;\;\;\; - expert for task $d$ \;\;\;\;\;\;\;\;\;\;\;\;\;\;\;\;\;\;\;\;\;
        $e_{s\rightarrow d}$  - NN edge from task $s$ to task $d$\\
        $P=\bigcup_{k=1}^{n_{iters}}part_{k}$ - dataset split \;\;\;\;\;\;\;\;\;\;\;\;\;\;\;\;\;\;\;\;\;\;\;\;\;\;\; T \;\;\;\; - the set of all tasks\\
        $\mathbf{W}_{i;d}$ - per-pixel Selection Ensemble weights, for sample $i$ and destination task $d$
    \algrule
    \textbf{Results:} 1) CShift node views $Y_{i;d}$; 2) all trained edges $e_{s\rightarrow d}$
    \begin{algorithmic}[1]
        \algrule
        \State $Y_{i;d} \gets \text{Expert}_d(X_{i}), \forall d \in T, \forall i \in P$ 
        \;\;\;\;\;\;\;\;\;\;
        \textcolor{algo_comments}{//  Fig.~\ref{fig:main_alg}-b) Generate the initial pseudo-labels}
        \For{$k \gets 1$ to $n_{iters}$} 
            \State $X_{i;s} \gets Y_{i;s}, \forall s\in T, \forall i \in P$ \;\;\;\;\;\;\;\;\;\;\;\;\;\;\; \textcolor{algo_comments}{// Update views - enable multi-level ensembles}
            \ForAll{$d \in T$} 
                \State train $e_{s\rightarrow d}(X_{i;s})=X_{i;d}, \forall i\in part_k, \forall s \in T$ 
                \textcolor{algo_comments}{// Fig.~\ref{fig:main_alg}-c1) Train with pseudo-labels}
                \State $Y_{i;d} \gets f(\bigcup\limits_{s}  \{e_{s\rightarrow d}(X_{i;s})\}\cup\{X_{i;d}\}, \mathbf{W}_{i;d}), \forall i \in P$ \;
                \textcolor{algo_comments}{ // Fig.~\ref{fig:main_alg}-c2) Selection ensemble }
            \EndFor
        \EndFor
    \end{algorithmic}
\end{algorithm}

\subsection{Multi-Task Graph}
\label{sec:approach_multi_task_graph}
We formally define the Multi-Task Graph over a set $T$ of tasks (\eg semantic segmentation, single-image depth estimation, surface normals - Sec.~\ref{sec:exp_analysis}), each illustrating a different view of the scene. There are one-to-one correspondences between graph nodes and the set of tasks, and each edge is an encoder-decoder neural net transformation between source and target task nodes (Fig.~\ref{fig:main_alg}-a). Consequently, our graph $G=(T,E)$ with $E=\{e_{s\rightarrow d}| e_{s\rightarrow d}(X_{s})=X_d, s,d\in T, s\neq d\}$, where $X_{s}$ is the scene representation under task $s$, and $e_{s\rightarrow d}$ is the neural network transforming the view between source task $s$ and destination task $d$. The graph edges are initialized using pre-trained expert teachers, one for each considered task.

\noindent\textbf{Passing an image through the graph:} given a raw $rgb$ frame, we associate it to the $rgb$ node. Next, we aim to enforce the consensual constraint in the graph: no matter what path the input $rgb$ takes through the graph, being transformed from one node to the next, it should have the same representation (view) at the same final node. 
    
\noindent\textbf{Initializing the graph:} Other approaches~\cite{ngc,xtc} start with a supervised training phase of all the out-edges of the $rgb$ node $\{e_{rgb\rightarrow d}| \forall d\in T\}$, requiring multi-task annotated datasets. As we work in the unsupervised regime for the target domain, our initial views (pseudo-labels) for different tasks are obtained from a set of out-of-distribution expert models (Fig.~\ref{fig:main_alg}-b). Each task node $d$ has an associated expert: $\text{Expert}_d$. Then the initial edges are trained by distilling the knowledge of the experts (see the list of experts in Sec.~\ref{sec:exp_analysis}). 

\subsection{Consensus Shift Learning}
\label{sec:approach_consensus_shift}
The consensus between edges reaching a given task $d$ provides a robust view for $d$. With each learning cycle, the node values (pseudo-labels) shift towards stronger consensus, following the CShift algorithm  (Alg.~\ref{alg:main_alg}). The new labels are then used to distill the single edges connecting them and thus set up the next learning stage. Each transformation is, in fact, the last step of a longer graph path, starting in the $rgb$ node and ending in a destination node $d$. All paths should ideally be in consensual agreement, but in practice, they are not, so an intelligent mechanism is needed to extract the robust knowledge shared by the majority. We employ the discovery of consensus among multiple outputs from incoming edges. Intuitively, CShift is an adaptive combination over the output of all edges reaching a destination. It is based on a similarity measure between those views, computed at pixel-level, which adaptively estimates the importance of each incoming edge, dynamically for each pixel and data sample. 

\begin{figure}[t]
    \centering
    \includegraphics[width=1\linewidth]{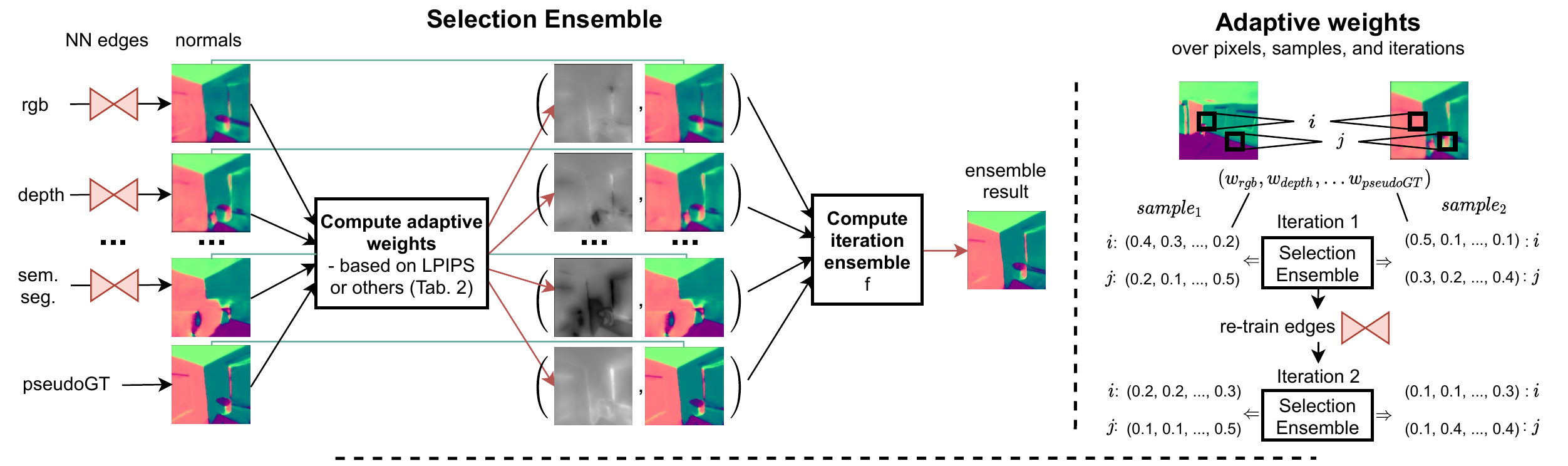}
    \includegraphics[width=1\linewidth]{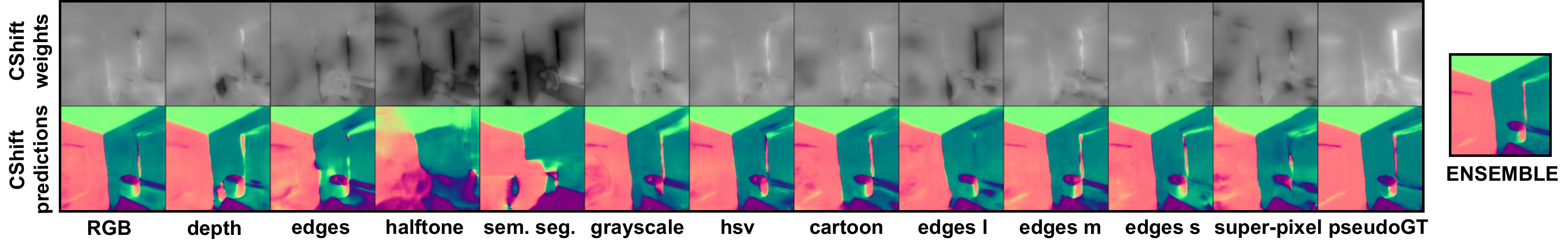}
    \setlength{\abovecaptionskip}{-20pt}
    \caption{ CShift selection flow, stripped down to pixel-level. The complex weights are highly adaptive, modifying, in effect, the underneath graph structure through their values. The last two rows show CShift per-pixel weights and the corresponding predictions from source tasks to $normals$ task destination. Note how the sharp zones correlate with high ensemble weights.}
    \label{fig:novelty_visual}
\end{figure}

Given a destination node $d$, all edges reaching this node $\{e_{s\rightarrow d}| s\in T, s\neq d\}$ are transformations from different views towards task $d$. For a sample $X_i$, CShift iteratively updates the sample's view, associated with task $d$, $X_{i;d}$. We define $\mathcal{N}(X_{i;d})$ to be the neighbourhood of $X_{i;d}$ as the set of all transformations from all different views of the sample joined with the current pseudo-label: $\mathcal{N}(X_{i;d}) = \{e_{s\rightarrow d}(X_{i;s}) | \forall s\in T\} \cup \{X_{i;d}\}$. The current task representation is replaced by the consensual one, computed as a function $f$ gathering information from all the neighbours of $X_{i;d}$, parameterized by pixel-level weights $\mathbf{W}_{i;d}\in\mathbb{R}^{h \times w \times |T|}$ ($(h, w)$ is the image size): $X_{i;d} \leftarrow f(\mathcal{N}(X_{i;d}); \mathbf{W}_{i;d})$, capturing the consensus between predictions. $\mathbf{W}_{i;d}$ has a channel associated with each task node, which indicates the similarity of the corresponding prediction with the current value of $d$. Without loss of generality, we assume $d$ is a single channel task. For a location $(x,y)$ and a given task $s$, the weights are computed as follows:
    \begin{equation}
        \mathbf{W}_{i;d}[x,y,s] = \frac{K(dist(e_{s\rightarrow d}(X_{i;s}), X_{i;d})[x,y])}{\sum_{Z\in\mathcal{N}(X_{i;d})}K(dist(Z, X_{i;d})[x,y])},
        \label{eq:kernel}
    \end{equation}
where $dist:\mathbb{R}^{h\times w} \rightarrow \mathbb{R}^{h \times w}$ is a distance function capturing the similarity between two different prediction maps and $K:\mathbb{R}\rightarrow\mathbb{R}$ is the kernel function that determines the weight of nearby points. The algorithm aims to identify the areas of the prediction maps that are perceptually similar and push them further in the ensemble while downgrading regions that seem to be noisy and uncorrelated with the other predictions. We propose a selection ensemble algorithm that automatically extracts the most representative consensual representation of the in-edges, being adaptive and changing separately for each pixel and data sample - essentially changing the graph's structure in a dynamic and per-pixel manner. The adaptive per-pixel weighting allows CShift to keep the most relevant information from all input maps, even when some maps are less reliable, treating similarities per region (kernels at pixel-level). In Sec.~\ref{sec:exp_analysis} we instantiate $dist, K, f$ and provide ablation experiments proving that the selection strategy is robust to noisy connections. We show in Fig.~\ref{fig:novelty_visual} how the selection based consensus works at pixel-level.

\section{Experimental Analysis}
\label{sec:exp_analysis}

\noindent\textbf{Datasets.} We perform experiments on Replica~\cite{replica} and Hypersim~\cite{hypersim}. Replica is a dataset of photo-realistic 3D indoor scenes, comprising $18$ scenes, with a total of $48$ rooms. In practice, we consider two iterations for training our Multi-Task Graph and use two unsupervised train sets ($9600$+$9600$ samples), a validation and a test set (each with $960$ samples). Hypersim is also a photorealistic synthetic dataset, for holistic indoor scene understanding. For two iteration training we use $10696$ + $10818$ samples, $319$ for validation and $1064$ for testing. During training and validation, only the raw $rgb$ images are available. We highlight that the annotations are only employed for evaluation purposes.

\noindent\textbf{Expert Models.} We employ per-task expert models, having only $rgb$ as input, trained on out-of-distribution data, to initialize the per-node pseudo-labels (Fig.~\ref{fig:main_alg}-b). First, the experts replace the direct edges starting from the $rgb$ node. Then, the computed views of a node become supervisory signals for the in-edges of that node and inputs for the out-edges (\eg for training edge $e_{depth\rightarrow halftone}$, the depth input is obtained by applying the depth expert over the $rgb$ frame and the halftone pseudo-label by extracting the halftone from $rgb$.) Our graph contains a total of $13$ task nodes, including $rgb$, thus we consider $12$ experts ranging from trivial color-space transformations to heavily trained deep nets for the following tasks: \textbf{1)} halftone, \textbf{2)} grayscale, \textbf{3)} hsv, \textbf{4)} depth, \textbf{5)} surface normals, \textbf{6, 7, 8)} small, medium and large scale low-level edges, \textbf{9)} high-level edges, \textbf{10)} super-pixel, \textbf{11)} cartoonization and \textbf{12)} semantic segmentation. The experts are trained on a wide variety of datasets, having a different distribution than ours. We detail the expert models in the supplementary material.

\noindent\textbf{Evaluation on three tasks.} Replica and Hypersim have similar annotation conventions only for $depth$ and $normals$ tasks. The XTC experts' considered for the tasks are not fully aligned with testing datasets annotations, as their training dataset uses different conventions. Thus, on $depth$, following the methodology of self-supervised methods~\cite{depth_align}, we performed a histogram specification alignment between expert results and ground truth annotations. On $normals$, we removed the 3rd channel in the XTC expert as Replica has $normals$ with only $2$ independent channels. We also report results for $rgb$, measuring the graph model's ability to reconstruct its original input through its many paths. We report the L1 error $\times 100$ (for readability).

\noindent\textbf{Implementation and training details.} Each graph edge is a neural network with a UNet architecture, as previously validated in NGC and XTC. The $156$ graph edges have $\approx4.3$ million parameters each, with $4$ down-scaling and $4$ up-scaling layers and a proper number of input and output channels, depending on the source and destination tasks. We optimize them by jointly minimizing $L2$ and the Structural Similarity Index Measure~\cite{ssim} (SSIM) losses for regression tasks equally weighted for each neural net. For training edges going to classification tasks (semantic segmentation or halftone), we use Cross-Entropy loss. As optimizer we work with SGD with Nesterov (lr=5e-2, wd=1e-3, momentum=0.9), and a ReduceLRonPlateau scheduler (patience=10, factor=$0.5$, threshold=1e-2, min lr=5e-5). We train $100$ epochs for the $1^\text{st}$ iteration and $100$ for the $2^\text{nd}$ one. Note that for the second iteration, we train all edges from scratch. 

\begin{figure}
    \begin{minipage}{0.49\columnwidth}
    \begingroup
        \setlength{\tabcolsep}{2.5pt} 
        \begin{tabular}{lccc} 
            \toprule
            \textbf{Method} & $depth$ & $normals$ & $rgb$ \\ 
            \toprule
            Expert \cite{xtc} & 14.58 & 8.30 & -\\
            Mean Ensemble \cite{ngc} & 12.94 & 7.95 & 4.30\\
            \midrule
            \textbf{CShift} w/ Variance & 12.80 & 7.91 & 2.12\\
            \textbf{CShift} w/ PSNR & 12.89 & 8.12 & 4.25 \\
            \textbf{CShift} w/ SSIM  & 12.80 &  7.89 & 2.38\\
            \textbf{CShift} w/ L1 & 12.81 & 7.73 & 2.16\\
            \textbf{CShift} w/ L2 & 12.79 & 7.72 & 2.45 \\
            \textbf{CShift} w/ LPIPS & \textbf{12.77} & \textbf{7.61} & \textbf{2.06} \\
            \bottomrule
        \end{tabular}
        \setlength{\abovecaptionskip}{0pt}
        \captionof{table}{Ablation study on different distance metrics on Replica dataset, for the first iteration. In all considered configurations CShift overcomes the initial expert models and in all, except the PSNR case, the Mean Ensemble.}
        \label{tab:ablation_ensembles}
    \endgroup
    \end{minipage}
    \hfill
    \begin{minipage}{0.49\columnwidth}
        \includegraphics[width=1\columnwidth, height=122pt]{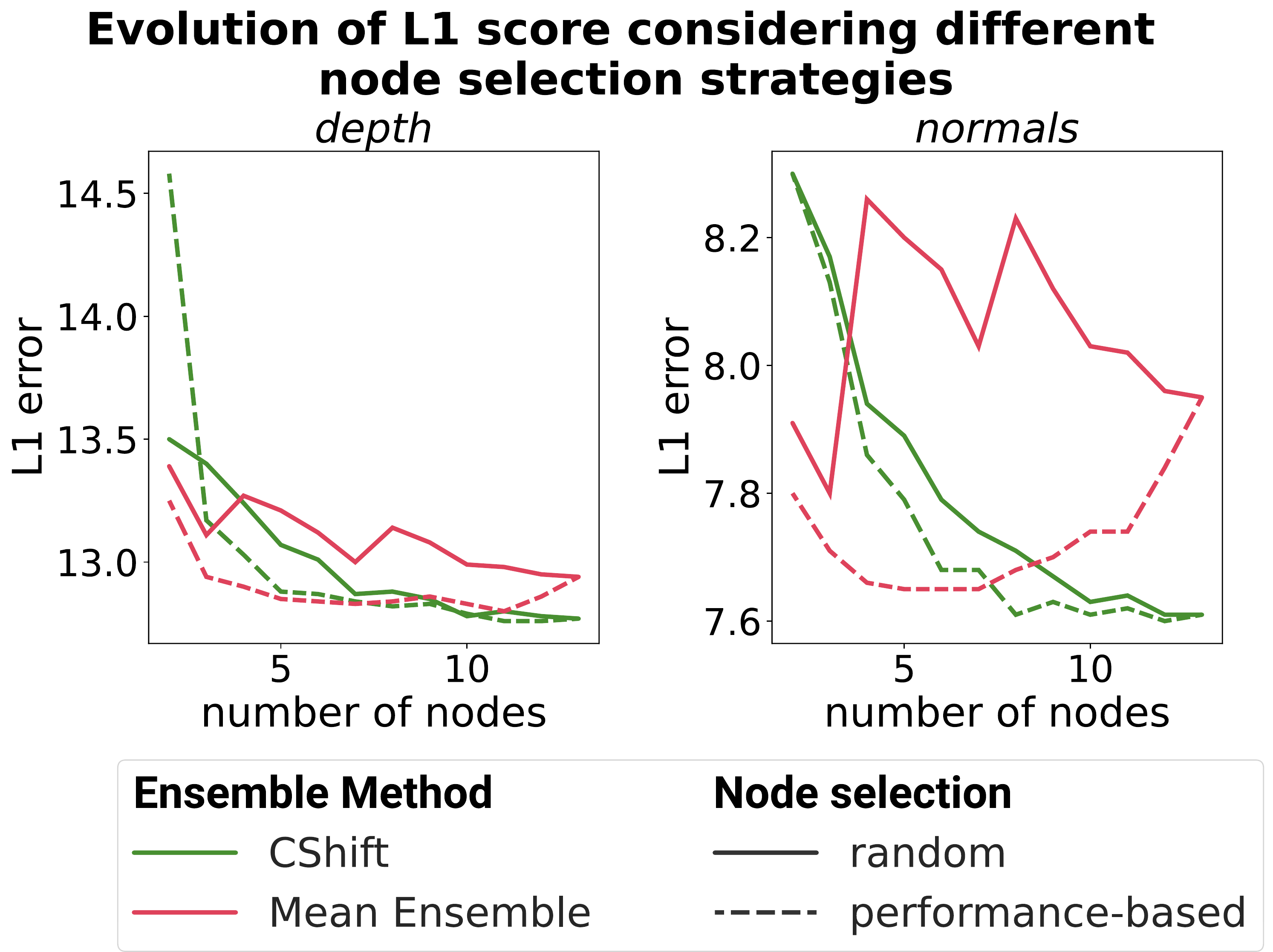}
        \setlength{\abovecaptionskip}{-25pt}
        \captionof{figure}{CShift is stable under different node selection strategies, improving even when using low-performing edges. In contrast, the mean ensemble has an unstable evolution under random node selection, its performance decreasing when weak edges are added.}
        \label{fig:plot_diff_n_nodes}
    \end{minipage}
\end{figure}

\noindent\textbf{Ensemble Selection Method.} At the core of the selection procedure is the distance function dictating each prediction map's per-pixel weights, as detailed in Sec.~\ref{sec:approach_consensus_shift}. We instantiate $f$ to the weighted median, and the kernel function $K$ (Eq.~\ref{eq:kernel}) to identity. To understand the selection strategy's power, we consider distance metrics ranging from local per-pixel distances to global perceptual measures: \textbf{1)} L1 and \textbf{2)} L2 distances at pixel-level \textbf{3)} Peak signal-to-noise ratio (PSNR) to measure the noise of the predictions; \textbf{4)} Structure Similarity Index Measure (SSIM)~\cite{ssim} that analyses the luminance, contrast and structural differences; \textbf{5)} Learned Perceptual Image Patch Similarity (LPIPS)~\cite{lpips} that is a deep model trained to identify perceptually similar images; \textbf{6)} per-pixel Variance among the multi-path predictions, to quantify their consensus. In Tab.~\ref{tab:ablation_ensembles} we compare $1^\text{st}$ CShift iteration under different selection strategies. Our proposed selections overcome the expert models and the simple mean ensemble in almost all the considered configurations (except for PSNR), highlighting the robustness of the process. Note that the simple mean ensemble baseline is similar to some extent with NGC. Our models are only trained using the expert models while NGC employs a supervised initialization step. We chose for all the subsequent experiments CShift w/ LPIPS.

\begin{figure}[t]
    \centering
    \includegraphics[width=1\linewidth]{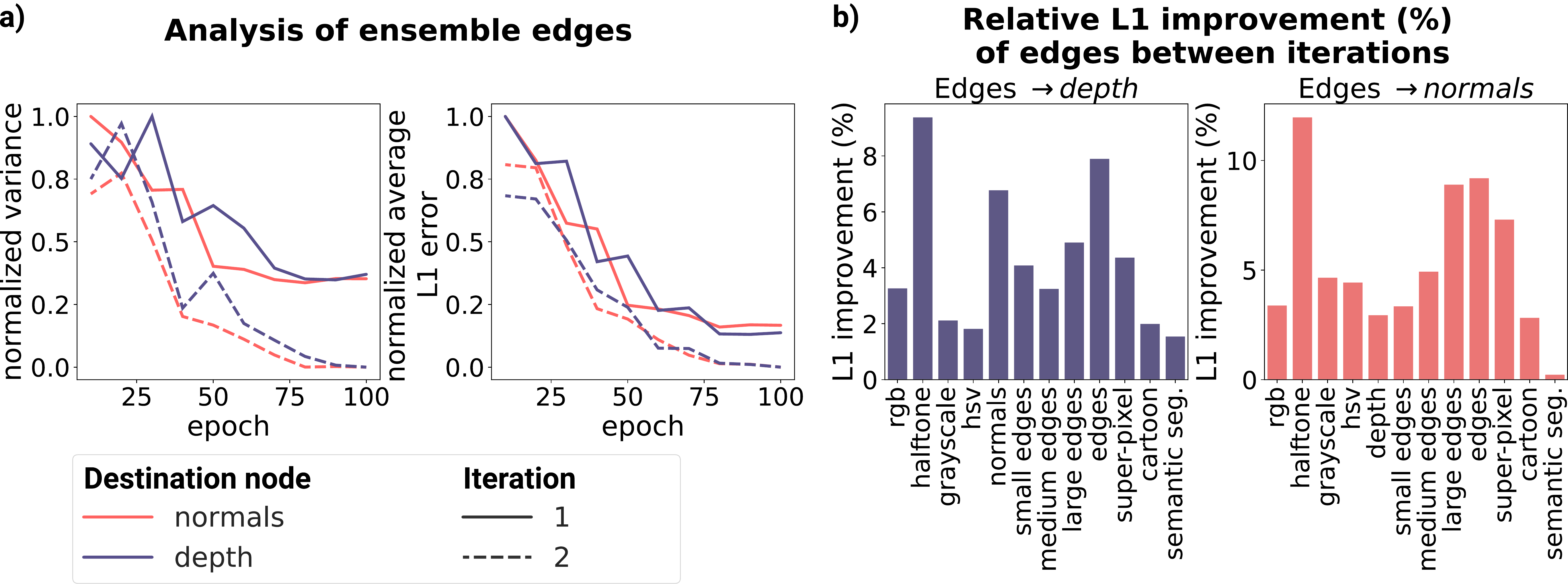}
    \setlength{\abovecaptionskip}{-25pt}
    \caption{\textbf{a)} The left plot shows that the average variance over single edges in an ensemble decreases over the training epochs and iterations. Thus, the graph's average consensus improves from one iteration to the next, while the average L1 error of edges decreases (right). CShift is effective and edges evolve towards the GT rather than collapsing in trivial solutions. \textbf{b)} We present the relative performance improvement of the individual edges between iterations. The performance of each edge increases up to almost $12$\%, proving the capacity of CShift to iteratively adapt to the new domain, in an unsupervised manner.}
    \label{fig:plot_var_errors_and_indiv_edges}
\end{figure}

\noindent\textbf{Consensus under different sets of nodes.} To validate that our model is robust to the set of the considered nodes (and their corresponding in and out edges), we perform an experiment where we start with a small graph containing only two nodes, and, step-by-step, increase the number of nodes until reaching all $13$ nodes. The nodes are added to the graph in a specific order. We analyze two ways of establishing this order: 1) random - nodes are randomly sorted; 2) performance-based - nodes are sorted according to their individual performance (evaluated w.r.t. to ground truth annotations) and added in this order. In Fig.~\ref{fig:plot_diff_n_nodes} we present the results of our experiment, comparing CShift with a mean ensemble baseline, for two destination tasks: $depth$ and $normals$. In both scenarios, the performance of CShift increases with the number of nodes, proving that our ensemble selection mechanism is able to extract relevant information even from low-performing edges. Note the performance fluctuations in baseline showing it is highly dependent on each edge reaching the ensemble.

\begin{figure}[t]
    \centering
    \includegraphics[width=1\linewidth]{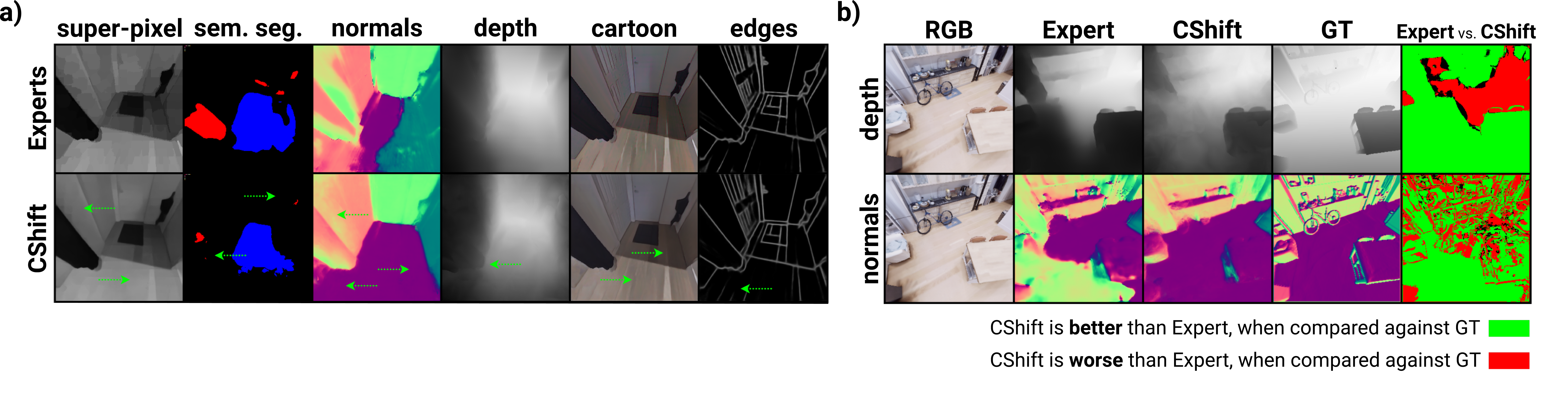}
    \setlength{\abovecaptionskip}{-30pt}
    \caption{\textbf{a)} We show the outputs of experts' used as pseudo-labels ($1^{st}$ row) and of CShift ($2^{nd}$ row). Green arrows point to significant improvements. From left to right are the representation for multiple tasks: super-pixel and cartoonization have less noise; semantic segmentation removes almost all pixels wrongly classified as ceiling; surface normals are significantly corrected; depth catches new details from the curtain; on edges, it removes some of the noisy edges coming from the floor texture. \textbf{b)} We compare the Expert with CShift (col 2-3). In the last column, we show the differences, green for pixels with improved prediction and red for weaker ones. CShift (second row) adds a bike in the scene, where the Expert completely misses it. The tasks are naturally interconnected and CShift takes advantage of that.}
    \label{fig:calitatives_all}
\end{figure}
\noindent\textbf{Edges improvement between iterations.} We give an in-depth analysis of how individual edges evolve over training epochs and CShift iterations. First, in Fig.~\ref{fig:plot_var_errors_and_indiv_edges}-a), we see how the variance between the edges in an ensemble decreases with more training. This is natural since each edge in the ensemble uses the same pseudo-GT. But, in the second iteration, the variance is even smaller, showing a smoother training optimization for the edges (which are trained from scratch for this second iteration). This could be explained by the new pseudo-labels coming from iteration 1 ensembles, rather than experts, making the training process simpler. To validate that the edges do not collapse to a bad representation, we also plot the average L1 error in Fig.~\ref{fig:plot_var_errors_and_indiv_edges}-a), confirming that all the edges improve their performance towards the GT. We show those relative improvements per edge, between the two CShift iterations in Fig.~\ref{fig:plot_var_errors_and_indiv_edges}-b).


\noindent\textbf{Qualitative views for multiple tasks.}  In Fig.~\ref{fig:calitatives_all}-a), we show the differences between the expert output, used as initial pseudo-ground truth for our graph edges (Fig.~\ref{fig:main_alg}-b), and the output of our CShift algorithm. Notice that the output of CShift looks smoother and partially corrects the mistakes of the expert model, adding significant value to the output.

\noindent\textbf{Qualitative results for tasks with GT.} We compare next in Fig.~\ref{fig:calitatives_all}-b) our results with the Expert, w.r.t. ground truth. Notice that CShift improves the output at a profound level, bringing in new information in the scene (see the bike in the second row). This is due to the multiple different source tasks for the ensembles' in-edges.

\begingroup
\setlength{\tabcolsep}{4pt} 
\begin{table}[t!]
    \setlength{\belowcaptionskip}{-4pt}
    \begin{center}
        \begin{tabular}{llrcr  rcr}
            \toprule
          & \multicolumn{1}{c}{\textbf{Method}} & \multicolumn{3}{c}{\shortstack{\textbf{Replica} $dest_{task}$ (L1 $\downarrow$)}} 
          & \multicolumn{3}{c}{\shortstack{\textbf{Hypersim} $dest_{task}$  (L1 $\downarrow$)}}\\
            \cmidrule(lr){3-5}
            \cmidrule(lr){6-8}
            & & $depth$ & $normals$ & $rgb$
            & $depth$ & $normals$ & $rgb$\\
            \toprule
            & Expert \cite{xtc} & 14.58 & 8.30 & - & 15.11 & {\color{blue}\textbf{9.10}} & -\\
            \midrule
            \parbox[t]{2mm}{\multirow{3}{*}{\rotatebox[origin=c]{90}{\textbf{Iter 1}}}} 
            & Average of direct edges & 14.32 & 9.34 & 6.33 & 16.91 & 12.55 & 11.34 \\
            & Edge: $rgb \rightarrow dest_{task}$ & 13.42 & 8.23 & - & 15.97 & 11.75 & - \\
            \cmidrule(l){2-8}
            & Mean Ensemble \cite{ngc} & 12.94 & 7.95 & 4.30  & 14.84 & 10.56 & 8.22 \\
            & \textbf{CShift} & 12.77 & {\color{red}\textbf{7.61}} & 2.06 & 13.98 & 9.36 & 3.56\\
            \midrule
            \parbox[t]{2mm}{\multirow{3}{*}{\rotatebox[origin=c]{90}{\textbf{Iter 2}}}} 
            &  Average of direct edges & 13.70 & 8.83 & 5.00 & 15.74 & 11.37 & 9.01 \\
            & Edge: $rgb \rightarrow dest_{task}$ & {\color{blue}\textbf{12.98}} & {\color{blue}\textbf{7.95}} & - & {\color{blue}\textbf{15.03}} & 10.09 & -\\
            \cmidrule(lr){2-8}
            & Mean Ensemble & 12.87 & 7.91 & 3.18 & 14.20 & 9.90 & 6.31\\
            & \textbf{CShift} &  {\color{red}\textbf{12.71}} & {\color{red} \textbf{7.61}} & {\color{red}\textbf{1.51}} & {\color{red} \textbf{13.75}} & {\color{red} \textbf{9.02}} & {\color{red} \textbf{1.84}}\\
            \midrule
            \midrule
            & \textbf{CShift} Boost $\uparrow$ & \textbf{12.8\%} & \textbf{8.3\%} & - & \textbf{9.0\%} & \textbf{0.9\%} & -\\
            \bottomrule
        \end{tabular}
    \end{center}
    \caption{Quantitative results. We compare our performance over each iteration against the initial experts, on destination tasks for which we have GT annotations. CShift ensemble outperforms XTC experts on $depth$ and $normals$ by a large margin, without any additional supervision. Even single, direct edges ($rgb \rightarrow dest_{task}$) improve over iterations, achieving better results compared with the experts in most of the cases (except for Hypersim's $normals$). With blue we represent the best single edge in the column and with red the best ensemble.}
    \label{tab:sota_comparison}
\end{table}
\endgroup

\noindent\textbf{Comparison with other methods.} Starting from pseudo-labels provided by the experts, we improve their quality by a large margin with our CShift ensemble and even with a direct link from $rgb$ (except for Hypersim's normals where GT is extremely detailed, while our single edge is a simple UNet with $4.3$ mil params). We achieve this performance (Tab.~\ref{tab:sota_comparison}) in just two CShift iterations, without adding any supervised information, largely outperforming the basic mean ensemble. Notice that the direct edges in the graph significantly improves over CShift iterations (in average and the individual direct edge from $rgb$). We treat all tasks unitary, so we tested the $rgb$ reconstruction performance, noticing large improvements over iterations.

\begin{figure}
    \hspace*{\fill}%
    \begin{minipage}{0.66\columnwidth}
    \begingroup
        \setlength{\tabcolsep}{2.5pt} 
        \begin{tabular}{l cccccc}
            \toprule
            \multicolumn{1}{|l|}{$dest_{task}$: $sem.\text{ }seg.$}& \multicolumn{6}{c}{\textbf{Expert models} (L1 $\downarrow$)}\\
            &\multicolumn{6}{c}{(from 1-weak to 6-strong) }\\
            \cmidrule(lr){2-7}
            \textbf{Method} & 1 & 2 & 3 & 4 & 5 & 6 \\
            \midrule
            Expert & 2.9 & 2.5 & 2.2 & 1.8 & 1.6 & 1.5\\
            \cmidrule(lr){1-7}
             \shortstack[l]{Mean Ensemble} & 2.6 & 2.5 & 2.3 & 2.1 & 1.9 & 2.0 \\
             \textbf{CShift} & 2.4 & 2.3 & 2.0 & 1.7 & 1.5 & 1.4\\
             \midrule
             \midrule
             \textbf{CShift} Boost $\uparrow$ & 15.7\% & 10.6\% & 8.2\% & 6.3\% & 6.1\% & 5.5\% \\
            \bottomrule
        \end{tabular}
        \setlength{\abovecaptionskip}{-10pt}
        \captionof{table}{The boost over the expert varies inversely with the expert strength (from left to right). As expected, it is harder to improve over a good expert, but CShift does it in all tests.}
        \label{tab:perf_ev_sem_seg_exp}
    \endgroup
    \end{minipage}
    \hfill
    \begin{minipage}{0.3\columnwidth}
    \includegraphics[width=1\columnwidth]{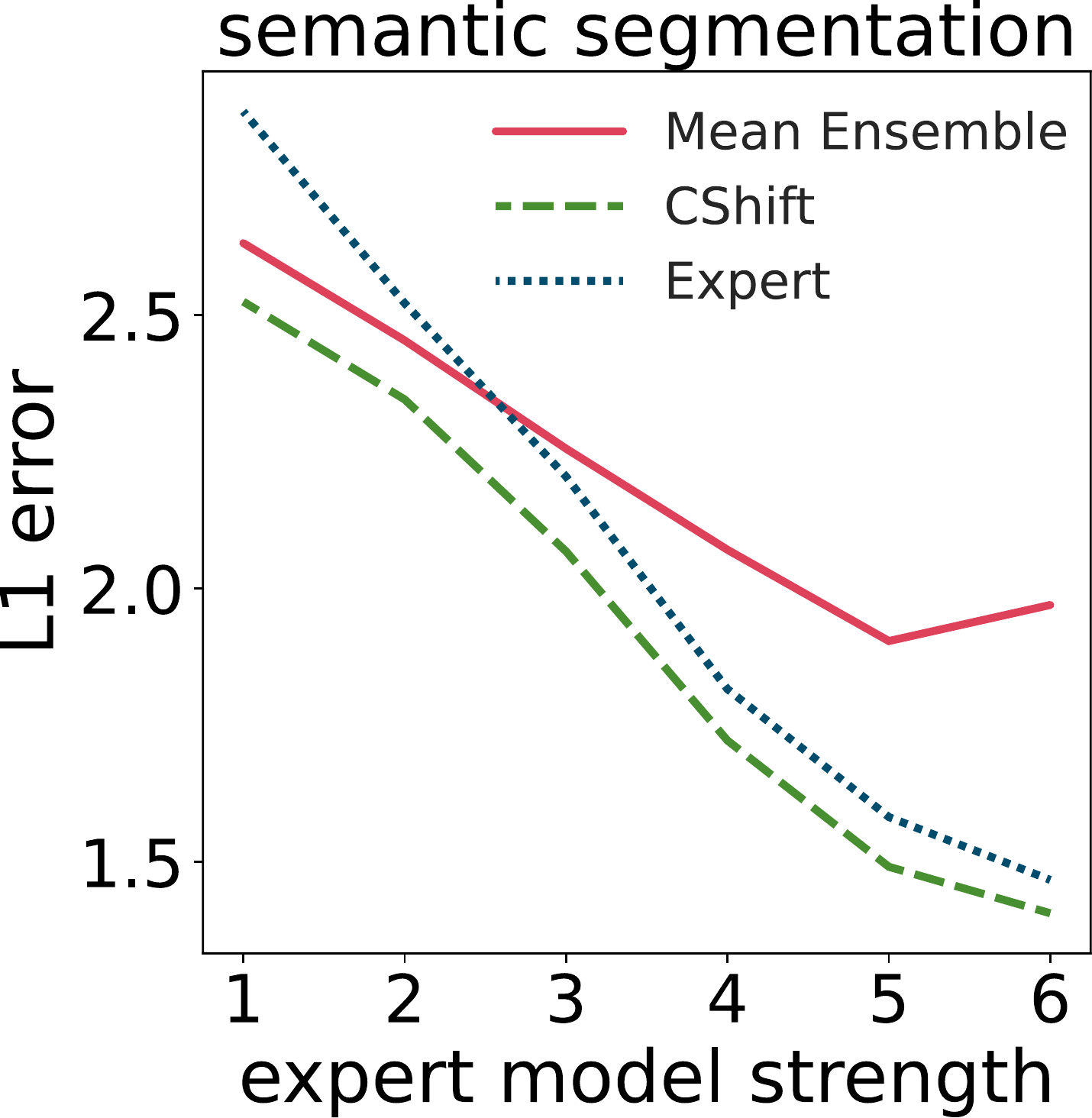}
        \setlength{\abovecaptionskip}{-20pt}
        \captionof{figure}{CShift consistently improves over experts of different quality. 
       }
        \label{fig:perf_ev_sem_seg_exp}
    \end{minipage}
\end{figure}

\noindent\textbf{Using weak expert models.} We test the importance of using weaker experts and expand the quantitatively validated domains set on semantic segmentation tasks (Hypersim dataset, with 40 classes). To control the level of expertise of our initial expert, we have trained it from scratch in a supervised manner on randomly chosen $10\%$, $30\%$ and $50\%$ of training samples of Hypersim, for 30 and 40 epochs, resulting in 6 expert models, with increasing performance. The relatively small training sets and the number of training epochs are specifically chosen to ensure that the experts are weak classifiers. This case is different and complementary to the other experiments in which we brought state-of-the-art experts pretrained on other datasets. Then we use these weak "expert" models, one at a time, as the regular experts in our first CShift iteration. We report L1 errors of class probability maps consistent with all the other tasks. In all cases, CShift outperforms both the initial expert and the baseline mean. A remarkable observation is that the gap between the expert and CShift varies inversely with the power of the expert: the weaker the expert, the higher the improvement (Tab.~\ref{tab:perf_ev_sem_seg_exp}, Fig.~\ref{fig:perf_ev_sem_seg_exp}).


\vspace{-10pt}

\section{Concluding remarks}
\label{sec:conclusion}

We introduce the CShift algorithm in multi-task graphs, able to learn unsupervised in new data distributions, using as supervision an intelligent consensus among the multiple pathways reaching a given task node, which adapts for each individual pixel and data sample. CShift's unsupervised capability and intelligent per-pixel ensemble selection for creating pseudo-labels make it significantly different and stronger than related methods. All key aspects of our approach, namely the CShift selection ensemble, the unsupervised domain adaptation capability, and the ability to learn from weak experts, are experimentally validated on two challenging datasets. Also, the comparisons to recent related works prove superior capabilities in the unsupervised learning case. We believe that CShift brings theoretically interesting and practically valuable contributions in an area of research, that of multi-task unsupervised learning, which is of major importance in today's machine learning.

\clearpage
\section*{Acknowledgements}
This work was funded in part by UEFISCDI, under Projects EEA-RO-2018-0496 and PN-III-P4-ID-PCE-2020-2819.

\begin{appendices}

In the following supplementary material, we provide additional details and experimental results that emphasize our main contributions. Sec.~\ref{sec:mmd} analyzes the distribution gap between the source domains of the expert models and our target domains. Sec.~\ref{sec:exp_models} provides additional details regarding the considered expert models.
 
\section{Out-of-distribution experts adaptation}\label{sec:mmd}
CShift requires no human-annotated data for the target domain. We take advantage of existing state-of-the-art expert models that distill research years and valuable expertise and provide reliable pseudo-labels for each of the considered tasks. When applied to novel domains, the weakness of these experts is that they are trained on different distributions. We first transfer their knowledge in our graph edges. Then our learning method, by exploiting and enforcing the overall consensus among all tasks, allows the graph to adapt by itself to the target domain, thus overcoming the domain gap, as shown in the following.

To emphasize the domain adaptation capabilities of CShift, we employ the Maximum Mean Discrepancy~\cite{mmd} (MMD) method for measuring the domain dissimilarity between our target domain and the expert source domains. MMD is a strong and widely used \cite{mmd_pami2021, mmd_icml2017,  mmd_cvpr2017} non-parametric metric for comparing the distributions of two datasets. We follow the methodology in \cite{mmd} and compute the unbiased empirical estimate of squared MMD. Our experiments show (Tab.~\ref{tab:mmd_1}) that there is a large distributional shift between our target domain and the domains of the original expert models. In conjunction with the ones presented in the Experimental Analysis Section of our main paper, these results prove our method's unsupervised domain adaptation capabilities.

We further analyze the gap between the source domain of $depth$ and $normals$ experts and one of our testing datasets: Replica. The experts \cite{xtc} are originally trained on Taskonomy dataset, which is a real-world dataset, while Replica is a synthetic dataset. We will compute the discrepancy in distribution using MMD as mentioned above. Considering that the obtained discrepancy is not an absolute measure, we will also use the synthetic Hypersim dataset to perform a relative comparison. The analysis is performed both for the input level and the expert's mid-level features. For computing MMD, we average over multiple runs, each containing 100-1600 samples per dataset. The results in Tab.~\ref{tab:mmd_1} show that there is a significant domain shift in the input for the pre-trained experts on Taskonomy, both at the $rgb$ level but also through the eyes of the experts ($depth$ and $normals$ columns). Notice that the Hypersim dataset is closer to Replica (compared with Taskonomy) since both use synthetic data.

\begin{table}[t]
\begin{center}
\begin{tabular}{l c c c}
\toprule
& $rgb$ & $depth$ & $normals$ \\
\toprule
MMD($\text{replica}_{part1}$, $\text{replica}_{part2}$) & 5.4 & 17.8 & 17.4 \\
MMD($\text{replica}_{part1}$, hypersim) & 3.4 & 20.1 & 20.6 \\
MMD($\text{replica}_{part1}$, taskonomy) & 13.1 & 23.3 & 20.2\\
\bottomrule
\end{tabular}
\end{center}
\caption{We report the MMD between one of our target domains (Replica dataset) and the source domain of the $depth$ and $normals$ expert models (Taskonomy dataset), considering both $rgb$ input and mid-level embeddings of the experts. Compared to another synthetic dataset (Hypersim),  we observe a smaller distribution shift than for Taskonomy, which contains real-world samples. We also validate our assumptions by comparing two different splits of Replica. For readability, we report MMD $\times 100$.}
\label{tab:mmd_1}
\end{table}

\section{Expert models}\label{sec:exp_models}
Our graph contains a total of $13$ task nodes, including the $rgb$ one, thus we consider $12$ experts ranging from trivial color-space transformations to heavily trained deep nets: \textbf{1)} halftone computed using \hyperlink{https://github.com/philgyford/python-halftone}{python-halftone}; \textbf{2)} grayscale and \textbf{3)} hsv computed with direct color-space transformations; \textbf{4)} depth and \textbf{5)} surface normals obtained from the XTC~\cite{xtc} experts; \textbf{6, 7, 8)} small, medium and large scale edges extracted using a Sobel-Feldman filter~\cite{sobel}, and more complex \textbf{9)} edges extracted using the DexiNed~\cite{dexined} expert; \textbf{10)} super-pixel maps extracted using SpixelNet~\cite{spixelnet}; \textbf{11)} cartoonization got from WBCartoon~\cite{cartoon_wb} and \textbf{12)} semantic segmentation maps computed with HRNet~\cite{hrnet}. The deep nets expert models are trained on a large variety of datasets: \textbf{4)} and \textbf{5)} Taskonomy~\cite{taskonomy}, \textbf{9)} BIPED~\cite{biped}, \textbf{10)} SceneFlow~\cite{sceneflow} + BSDS500~\cite{BSDS500}, \textbf{11)} FFHQ~\cite{ffhq}, \textbf{12)} ADE20k~\cite{ade20k}. Note that these datasets are built for a different purpose, on a different distribution than ours. 
\end{appendices}

\bibliography{egbib}
\end{document}